\documentclass{article}

\newif\ifpreprint%
\preprinttrue   %

\PassOptionsToPackage{numbers,square}{natbib}
\usepackage{natbib}
\ifpreprint
\usepackage{newpxtext,newpxmath}
\usepackage[left=1.5in,right=1.5in]{geometry}
\else
\usepackage[preprint]{neurips_2021}
\fi

\newlength{\mywidth}
\ifpreprint
\else
\fi

\usepackage{microtype}
\usepackage{graphicx}
\usepackage{subfigure}
\usepackage{booktabs} %
\usepackage{amssymb}
\usepackage{amsmath}
\usepackage{setspace}
\usepackage{tikz}
\usetikzlibrary{positioning,arrows.meta,matrix,fit}
\usepackage{gnuplot-lua-tikz}
\pgfdeclarelayer{background}
\pgfdeclarelayer{foreground}
\pgfsetlayers{background,main,foreground}
\usepackage{balance}
\usepackage{supertabular}
\usepackage[noend]{algorithmic}
\usepackage{algorithm}
\usepackage{enumitem}

\usepackage[utf8]{inputenc} %
\usepackage[T1]{fontenc}    %
\usepackage{hyperref}       %
\usepackage{url}            %
\usepackage{amsfonts}       %
\usepackage{nicefrac}       %
\usepackage{microtype}      %
\usepackage{xcolor}         %

\newcommand{\diag}{\mathop{\mathrm{diag}}}

\newcommand{\dotdot}{\mathrel{{.}\,{.}}\nobreak}

\ifpreprint
\title{Accelerating Quadratic Optimization\\with Reinforcement Learning}
\else
\title{Accelerating Quadratic Optimization with Reinforcement Learning}
\fi

\ifpreprint
\author{\hspace{-.5em}Jeffrey Ichnowski$^{{\mbox{\large*}}1}$, Paras Jain$^{{\mbox{\large*}}1}$, Bartolomeo Stellato$^2$,\\
Goran Banjac$^3$, Michael Luo$^1$, Francesco Borrelli$^1$,\\ Joseph E. Gonzalez$^1$, Ion Stoica$^1$, and Ken Goldberg$^1$ \\[1em]
$^1$University of California, Berkeley \\
$^2$Princeton University \\
$^3$ETH Zurich}
\date{}
\else
\author{Jeffrey Ichnowski$^1$, Paras Jain$^1$, Bartolomeo Stellato$^2$, Goran Banjac$^3$, Michael Luo$^1$, \\ Francesco Borrelli$^1$, Joseph E. Gonzalez$^1$, Ion Stoica$^1$, and Ken Goldberg$^1$ \\
$^1$University of California, Berkeley \\
$^2$Princeton University \\
$^3$ETH Zurich
}
\fi

\newcommand{\ours}{RLQP}

\begin{document}

\maketitle

\begin{abstract}

First-order methods for quadratic optimization such as OSQP are widely used for large-scale machine learning and embedded optimal control, where many related problems must be rapidly solved. These methods face two persistent challenges: manual hyperparameter tuning and convergence time to high-accuracy solutions. To address these, we explore how Reinforcement Learning (RL) can learn a policy to tune parameters to accelerate convergence. In experiments with well-known QP benchmarks we find that our RL policy, RLQP, significantly outperforms state-of-the-art QP solvers by up to 3x. RLQP generalizes surprisingly well to previously unseen problems with varying dimension and structure from different applications, including the QPLIB, Netlib LP and Maros-M{\'e}sz{\'a}ros problems.
Code for RLQP is available at \url{https://github.com/berkeleyautomation/rlqp}.

\end{abstract}

{\let\thefootnote\relax\footnotetext{{$^*$ equal contribution}}}
\section{Introduction}
\label{sec:introduction}

Solving quadratic programs (QPs) efficiently is critical to applications in finance, robotic control and operations research.
While state-of-the-art interior-point methods scale poorly with problem dimensions, first-order methods for solving QPs typically require thousands of iterations.
Moreover, real-time control applications have tight latency constraints for solvers~\cite{mattingley2012cvxgen}.
Therefore, it is important to develop efficient heuristics to solve QPs in fewer iterations.

The Alternating Direction Method of Multipliers (ADMM)~\cite{boyd2011distributed,gabay1976dual,glowinski1975approximation} is an efficient first-order optimization algorithm, and is the basis for %
the widely used and state-of-the art Operator-Splitting QP (OSQP) solver~\cite{osqp}. ADMM performs a linear solve on a matrix based on the optimality conditions of the QP to generate a step direction, and then projects the step onto the constraint bounds.

While state-of-the-art, the ADMM algorithm has numerous hyperparameters that must be tuned with heuristics to regularize and control optimization.
Most importantly, the step size parameter $\rho$ has considerable impact on the convergence rate. However, is still unclear how to select $\rho$ before attempting the QP solution. While some theoretical works compute the optimal~$\rho$~\cite{metric_selection_admm}, they rely on solving semidefinite optimization problems which are much harder than solving the QP itself.
Alternatively, some heuristics introduce ``feedback'' by adapting $\rho$ throughout optimization in order to balance primal and dual residuals~\cite{osqp,boyd2011distributed,he_alternating_2000}.

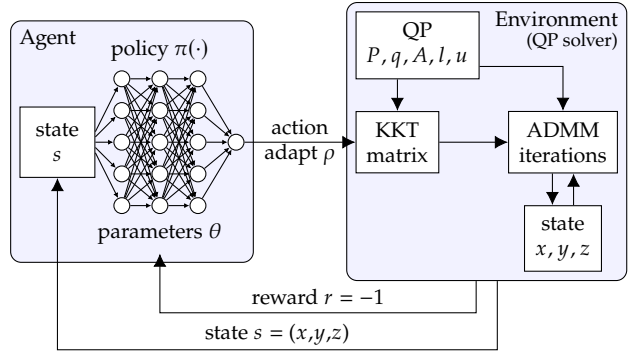
\begin{figure}
    \centering
    \begin{tikzpicture}[ %
  x=1pt,y=1pt, %
  every node/.style={font=\footnotesize},
  block/.style={rounded corners, fill=blue!5, draw, minimum height=32pt},
  iblk/.style={draw, align=center, inner sep=4pt, fill=white},
  neuron/.style={circle,draw,minimum size=6pt, inner sep=0pt, fill=white},
  >={Latex[width=5pt,length=5pt]},
  nn/.style={->,>={Latex[width=2pt,length=2pt]}}
]

\node [draw, align=center, inner sep=6pt, fill=white] (s) {
state \\
$s$
};

\foreach \y in {-2,-1,0,1,2} {
   \node [neuron,xshift=10pt,yshift=\y*12] (input-\y) at (s.east) {};
   \node [neuron,right=8pt of input-\y] (hidden-\y) {};
   \node [neuron,right=8pt of hidden-\y] (output-\y) {};
   \draw [nn] (s.east) ++ (0,\y*2) -- (input-\y);
}

\node [above=0pt of hidden-2] (network) { policy $\pi(\cdot)$ };
\node [neuron, right=8pt of output-0] (policy) {};
\node [below=0pt of hidden--2] (params) { parameters $\theta$};

\foreach \y in {-2,-1,0,1,2} {
  \foreach \z in {-2,-1,0,1,2} {
    \draw [nn] (input-\y) -- (hidden-\z);
    \draw [nn] (hidden-\y) -- (output-\z);
  }
  \draw [nn] (output-\y) -- (policy);
}

\node [iblk, right = 42pt of policy] (kkt) { KKT \\ matrix};

\node [iblk, above = 12pt of kkt.north west, anchor=south west] (qp) { QP \\
$P, q, A, l, u$};

\node [iblk, right = 26pt of kkt] (admm) { ADMM \\ iterations };

\node [iblk, below = 12pt of admm] (qpstate) {
state \\
$x,y,z$
};

\draw [<-] (kkt.north) -- ++(0,12pt);

\draw [->] (qp.east) ++ (0,-8pt) -| (admm);
\draw [->] (kkt) -- (admm);
\draw [->] ([xshift=-4pt]admm.south) -- ([xshift=-4pt]qpstate.north);
\draw [->] ([xshift=4pt]qpstate.north) -| ([xshift=4pt]admm.south);

\begin{pgfonlayer}{background}
  \node [block, fit=(s)(policy)(network)(params)] (agent-box) {};
  \node [below right=0pt of agent-box.north west] {Agent};
  \node [block, fit=(qp)(kkt)(admm)(qpstate)] (env) { }; %
  \node [below left=0pt of env.north east, align=right] (env-label) { Environment \\[-2pt] \scriptsize (QP solver) };
  \draw [->] (env.south) ++(4pt,0) -- ++(0,-26pt) -| node [near start, yshift = 6pt] {state $s = $ ($x$,$y$,$z$)} (s.south);
  \draw [->] (env.south) ++(-4pt,0) -- ++(0,-12pt) -| node [near start, yshift = 6pt] {reward $r = -1$} (params.south);
  \draw [->] (policy) -- node [yshift = 0pt, align=center] {action \\ adapt $\rho$ } (kkt);
\end{pgfonlayer}
\end{tikzpicture}
    \caption{RLQP uses deep reinforcement learning (RL) to compute a policy that adapts the internal parameters of a first-order quadratic program (QP) solver to speed up the solver's convergence rate.  In a standard RL formulation, a policy computes an action based on its observation of the state of the environment, and taking the action results in a change in state and a reward.  In RLQP, the policy is parameterized by a neural network, the state is the internal state of the QP solver, the action changes a parameter ($\rho$) of the solver, and the reward minimizes the time required to solve the QP.}
    \label{fig:rlqp}
\end{figure}

We propose RLQP (see Fig.~\ref{fig:rlqp}), an accelerated QP solver based on OSQP that uses reinforcement learning to adapt the internal parameters of the ADMM algorithm between iterations to minimize solve times. %
An RL algorithm learns a policy $\pi_\theta\colon\mathcal{S}\to\mathcal{A}$, parameterized by $\theta$ (e.g., the weights of a neural network), that maps states in a set $\mathcal{S}$ to actions in set $\mathcal{A}$ such that the selected action maximizes an accumulated reward $r$.
To train the policy for RLQP, we define $\mathcal{S}$ to be the internal state of the QP solver (e.g., the constraint bounds, the primal and dual estimates), $\mathcal{A}$ to be the adaptation to the internal parameter ($\rho$) vector, and $r$ to minimize the number of ADMM iterations taken.

RLQP's policy can be trained either jointly across general classes of QPs or with respect to a specific class. The general version of RLQP is trained once on a broad class of QPs and can be used out-of-the-box on new problems.
The specialized version of RLQP is trained on a specific class of problems that the solver will repeatedly encounter.  While this requires additional setup and training time, it is useful when QPs will be repeatedly solved in application (e.g., in a 100~Hz control loop).

In experiments, we train RLQP on a set of randomized QPs, and compare convergence rates of RLQP to non-adaptive %
and heuristic adaptive policies.  To compare generalization and specialization, we investigate RLQP's performance in the settings where 1) the train and test sets of QPs come from the same class of problems, 2) the train set contains from superset of classes contained in the test set, 3) the train set contains a subset, and 4) when the train and test sets are from distinct classes.  In the results section we show that RLQP outperforms OSQP by up to 3x.

The contributions of this paper are:
\begin{itemize}[leftmargin=0.25in]
    \item Two RL formulations to train policies that provide coarse (scalar) and fine (vector) grain updates to the internal parameters of a QP solver for faster convergence times
    \item Policies trained jointly across QP problem classes or to specialize to specific classes
    \item Experimental results showing that RLQP reduces convergence times by up to 3x and generalizes to different problem classes and outperform existing methods
\end{itemize}
\section{Related Work}
\label{sec:related_work}

This work touches a number of related research areas, including convex optimization, using machine learning (ML) to speed up optimization, learning in first-order methods, and reinforcement learning.

\paragraph{Convex optimization} Many researchers have proposed algorithms for quadratic programs, which generally fall into three classes: active set~\cite{wolfe1959simplex}, interior point~\cite{nesterov1994interior}, and first-order methods.
Of the active set and interior point solvers, perhaps the most well-known are Gurobi~\cite{Gurobi} and MOSEK~\cite{Mosek}.
Active-set solvers operate by iteratively adapting an active set of constraints based on the cost function gradient and dual variables~\cite{nocedal2006numerical}.
Interior-point solvers iteratively introduce and vary barrier functions to represent constraints and solve unconstrained convex problems.
We instead base this work on a first-order method solver, OSQP~\cite{osqp}.
One of the advantages of OSQP over interior points solvers, is that they can readily be warm started from a near-by solution, as is common in many applications such as solving a sequential quadratic program%
~\cite{schulman2013finding}
and solving QPs for model-predictive control.

\paragraph{ML-accelerated combinatorial optimization} Accelerating combinatorial optimization problems with deep learning has been explored with wide application~\cite{bengio2020machinesurvey, bertsimas2020voice}, including branch-and-bound for mixed-integer linear programming~\cite{balcan2018learning, khalil2016learning}, graph algorithms~\cite{dai2017learning} and boolean satisfiability problems (SAT)~\cite{chen2018learning}. Many combinatorial optimization problems have exponential search spaces and are NP-hard in a general setting. However, learning-augmented combinatorial algorithms utilize very different methods to \ours{} as combinatorial problems have discrete search spaces.

\paragraph{Learning in first-order methods} Accelerating first-order methods with machine learning has gained considerable recent interest. \citet{li2016learning} demonstrate a learned optimization algorithm outperforms common first-order methods for several convex problems and a small non-convex problem. \citet{metz2019understanding} show a learned policy outperforms first-order methods when optimizing neural networks, but finds that directly learning parameter update values can be sensitive to exploding gradient problems. We avoid this instability during optimization by learning a policy to adapt parameters of the ADMM algorithm. \citet{wei2020tuning} recently proposed an RL agent to tune parameters for an ADMM-based inverse imaging solver.

\paragraph{Reinforcement Learning Overview}
Reinforcement learning (RL) algorithms include both on-policy algorithms, such as Proximal Policy Optimization~\cite{schulman2017proximal}, REINFORCE~\cite{sutton1999policy}, and IMPALA~\cite{espeholt2018impala}, and off-policy algorithms, such as DQN~\cite{mnih2013playing} and Soft Actor Critic~\cite{haarnoja2018soft}.
\ours{} extends the off-policy Twin-Delayed DDPG (TD3)~\cite{fujimoto2018addressing}, 
an actor-critic framework with a exploration policy for continuous action spaces that extends Deep Deterministic Policy Gradient (DDPG) algorithm~\cite{lillicrap2015continuous} while addressing approximation errors.
Furthermore, in one formulation of \ours{}, we train a shared policy for multiple agents following an RL approach proposed by \citet{huang2020one}.  With this single policy, RLQP updates multiple parameters using state associated with each constraint of a QP.

\section{Background}
\label{sec:background}

In this section, we summarize QPs, the OSQP solver, and a MDP formalization.

\subsection{Quadratic Programs}

A quadratic program with $n$ variables and $m$ constraints takes the form:
\begin{equation*}
\begin{array}{ll}
\mbox{minimize} & (1/2)x^TPx + q^Tx\\
\mbox{subject to} & l \le Ax \le u,
\end{array} 
\end{equation*}
where ${x} \in \mathbb{R}^n$ is the optimization variable, ${P}$ is an $n \times n$ symmetric positive semi-definite matrix that defines the quadratic cost, ${q} \in \mathbb{R}^n$ defines the linear cost, ${A}$ is an $m \times n$ matrix that defines the $m$ linear constraints, and ${l},{u} \in \mathbb{R}^m$ are the constraint's lower and upper bounds.
Here, $\le$ is an element-wise less-than-or-equal-to operator.
In this form, to specify an equality constraint, the lower and upper bounds are set to the same value, and to specify a constraint unbounded from one side, a sufficiently large value (or $\pm\infty$) is specified for the other side.

\subsection{First-Order QP Solver Algorithm}

The solver we speed up is OSQP, which uses a first-order ADMM method to solve QPs.  We summarize OSQP here.  Given a QP, OSQP first forms a \emph{KKT} matrix (below), then iteratively refines a solution from a initialization point for vectors ${x}^{(0)} \in \mathbb{R}^n$, ${y}^{(0)} \in \mathbb{R}^m$, and ${z}^{(0)} \in \mathbb{R}^m$, where the superscript in parenthesis refers to the iteration.
Each iteration computes the values for the $k+1$ iterates by solving following linear system (e.g., with an $LDL^T$ solver):
\begin{equation}
\setstretch{0.9}
\label{eqn:system}
\underbrace{\begin{bmatrix}
P + \sigma I %
   & A^T \\
A & \hspace{-6pt}\diag(\rho)^{-1} %
\end{bmatrix}}_{\text{KKT matrix}}
\begin{bmatrix}
x^{(k+1)} \\
v^{(k+1)}
\end{bmatrix}\!
=\!
\begin{bmatrix}
\sigma x^{(k)} - q \\
z^{(k)} - \diag(\rho)^{-1}y^{(k)}
\end{bmatrix}
\end{equation}
and then performing the following updates:
\begingroup
\allowdisplaybreaks
\begin{align*}
\tilde{{z}}^{(k+1)} &\gets {z}^{(k)} + \diag(\rho)^{-1}({v}^{(k+1)} - {y}^{(k)}) \\
z^{(k+1)} &\gets \Pi \left(\tilde{z}^{(k+1)} + \diag(\rho)^{-1}y^{(k)} \right) \\
y^{(k+1)} &\gets x^{(k)} + \diag(\rho) \left( \tilde{z}^{(k+1)} - z^{(k+1)} \right),
\end{align*}
\endgroup
where $\sigma \in \mathbb{R}_+$ and $\rho \in \mathbb{R}_+^m$ are regularization and step-size parameters, %
and $\Pi : \mathbb{R}^m \rightarrow \mathbb{R}^m$ projects its argument on the constraint bounds.
We use the notation $\diag\colon\mathbb{R}^m\to\mathbb{S}^m$ to denote the operator that maps a vector to a diagonal matrix.
We define the primal and the dual residual vectors as 
\begin{align*}
    \xi_{\rm primal}^{(k)} = Ax^{(k)} - b,\quad \mbox{and}\quad \xi_{\rm dual}^{(k)} = Px^{(k)} + q + A^Ty^{(k)}.
\end{align*}
When the primal and dual residual vectors are small enough in norm
after $k$ iterations,
$x^{(k+1)}$ and $y^{(k+1)}$ are primal and dual (approximate) solutions to the QP.

Internally, OSQP has a single scalar $\bar\rho$ that it uses to form $\rho$ according to the following formula:
\begin{equation}
\label{eqn:rhobartorho}
\rho_i = \begin{cases}
\bar\rho & \text{if } l_i \ne u_i \text{ (inequality constraints)} \\
\bar\rho \cdot 10^3 & \text{if } l_i = u_i \text{ (equality constraints)},
\end{cases}
\end{equation}
where the subscript $i$ denotes the $i$-th coefficient of $\rho$, and the bounds ${l}$ and ${u}$.

Periodically, between ADMM iterations, OSQP will adapt the value of $\bar\rho$.
The existing hand-crafted formula for adapting $\rho$ attempts to balance between primal and dual residuals, by setting $\bar\rho^{(k+1)} \leftarrow \bar\rho^{(k)} \sqrt{\|\xi_\mathrm{primal}\| / \|\xi_\mathrm{dual}\|}$. %
Empirically, adapting $\rho$ between iterations can speed up the convergence rate. %

\subsection{Multi-Agent Single-Policy MDP}
\label{sec:mdp}
In a %
Markov Decision Process (MDP), an \emph{agent} can be in any state $s \in \mathcal{S}$, take an action $a \in \mathcal{A}$, and with the transition dynamics function, $\mathcal{T}(\cdot \mid s, a)$, transitions from state $s$ to state $s'$ after taking action $a$.  The agent receives a reward $R \colon \mathcal{S} \times \mathcal{A} \to \mathbb{R}$ for transitioning from $s$ to $s'$ by taking action $a$. %
Given a tuple $(\mathcal{S}, \mathcal{A}, T, R, %
\gamma)$, the %
MDP optimization objective is to find a policy $\pi_\theta : \mathcal{S} \rightarrow \mathcal{A}$, parameterized by $\theta$, that maximizes the expected cumulative reward
$
E \left[
\sum_{t=0}^{\infty} \gamma^t r^t
\right],
$
where $r^t$ is the reward at time $t$ and $\gamma \in [0,1)$ is a discount factor.

We also formulate a multi-agent single-policy MDP setting in which $m$ agents collaborate in a shared environment in state $s_\mathrm{env} \in \mathcal{S}_\mathrm{env}$. At each time step, each collaborating agent (CA) $i$ has its own state $s_i \in \mathcal{S}_\mathrm{ca}$, action $a_i \in \mathcal{A}_\mathrm{ca}$, and observations $o_i \in \mathcal{O}$, but, for computation feasibility, share a single policy $\pi_\theta : \mathcal{S}_\mathrm{ca} \rightarrow \mathcal{A}_\mathrm{ca}$.  State transitions for the environment and all $m$ agents occur simultaneously according to a state transition function $\mathcal{T} : \mathcal{S}_\mathrm{env} \times \mathcal{S}_\mathrm{ca}^{m} \times \mathcal{A}_\mathrm{ca}^m \rightarrow \mathcal{S}_\mathrm{env} \times \mathcal{S_\mathrm{ca}}^{m}$ and result in a single shared reward $R : \mathcal{S}_\mathrm{env} \times \mathcal{S}_\mathrm{ca}^m \times \mathcal{A}_\mathrm{ca}^m \rightarrow \mathbb{R}$ and discount factor.  The objective is to find a single shared policy $\pi_\theta$ that maximizes the expected cumulative reward. This can be thought of as a special case of a multi-agent MDP~\cite{lowe2017multi} or Markov game~\cite{littman1994markov}, and we adapt a formulation from Huang et al.~\cite{huang2020one}. %

\section{Method}
\label{sec:method}

The goal of \ours{} is to learn a policy to adapt the $\rho \in \mathbb{R}^m$ vector used in the ADMM update in~\eqref{eqn:system} (see Fig.~\ref{fig:rlqp}).  As the dimensions of this vector vary between QPs, we propose two methods that can handle the variation in $m$.  The first method learns a policy to adapt a scalar $\bar\rho$ and then applies~\eqref{eqn:rhobartorho} to populate the coefficients of the $\rho$ vector.  The second method %
learns a policy to adapt individual coefficients of the $\rho$ vector.

Since both the number of variables $n$ and the number of constraints $m$ can vary from problem to problem, and the same QP can be written in $(n!\,m!)$ permutations, we propose learning policies that are problem size and permutation invariant.  To do this, we provide a permutation-invariant fixed-size %
state of the QP solver to either policy. %

\subsection{RL Policy for Scalar Adaptation}

\begin{figure}[t]
\begin{minipage}{2.6875in}
\input{algorithm/ddpg}
\end{minipage}\hfill%
\begin{minipage}{2.6875in}
\input{algorithm/ddpg_vec}
\end{minipage}
\end{figure}

To speed up convergence of OSQP, we hypothesize that RL can learn a scalar $\bar\rho$ adaptation policy that can perform as-well-as or better than the current handcrafted policy ($\pi_\mathrm{hc}$) of OSQP.
The handcrafted policy in OSQP periodically adapts $\rho$ by computing a single scalar $\bar\rho$, then sets the coefficients of $\rho$ based on the value of $\bar\rho$.
In both handcrafted and RL cases, the policy is a function $\pi : S_{\bar\rho} \rightarrow A_{\bar\rho}$, where $S_{\bar\rho} \in \mathbb{R}^2$ are the primal and dual residuals stacked into a vector, $A_{\bar\rho} \in \mathbb{R}$ is the value to set to $\bar\rho$.
One advantage of this approach is that a simple heuristic can check that the proposed change to $\bar\rho$ is sufficiently small and avoid a costly matrix factorization.

To compute this policy, $\pi$, we use Twin-Delayed DDPG 
TD3~\cite{fujimoto2018addressing}, an extension of 
deep-deterministic policy gradients (DDPG)~\cite{lillicrap2015continuous}, as the action space is continuous. %
We summarize TD3 in Alg.~\ref{alg:ddpg}.
TD3 learns the parameters $\theta$ of a policy $\pi_\theta$ network and critic $Q$ network, where $\pi$ determines the action to take and $Q(s,a) = \mathbb{E}_{s'}[r(s,a) + \gamma \mathbb E_{a'\sim\pi}[Q(s',a')]]$ is the expected reward for a given state-action pair following the recursive Bellman equation.  TD3 updates $Q$ by minimizing the loss on the Bellman equation, and updates the policy network using a policy gradient~\cite{sutton1999policy} of the objective
\[
J(\theta) = \mathbb E_{s\sim\pi}[R(s,a)],
\]
that is,
\[
\nabla_\theta J = \mathbb E_{s\sim \mathcal{D}}[\nabla_\theta\pi_\theta(s) \nabla_a Q(s,a) |_{a=\pi(s)}]
\]
where $\mathcal{D}$ is the discounted state visitation distribution~\cite{silver2014deterministic}.
For brevity, we leave out some details of TD3 in the algorithms, including: $Q$ is composed of two networks, the minimum value of the two networks estimates the reward, exploration noise is clamped, and $\pi$ network updates are staggered.

In RLQP, the ``environment'' $\mathtt{env}$ is an instance of a randomized QP problem, and a call to $\mathtt{step}()$ applies a change to $\bar\rho$ (and thus via Eq.~\ref{eqn:rhobartorho} to $\rho$), advances a QP a fixed number of ADMM iterations, and returns the updated internal state $s$, a reward $r$, and a termination flag $\mathtt{done}$.
In this case, the internal state $s$ is a vector containing the current primal and dual residuals of the QP.
The reward $r$ is $-1$ if not done, and $0$ if the QP is solved.

We train with randomized QPs across various problem classes (Sec.~\ref{sec:experiments}) that have solutions guaranteed by construction. To ensure progress, we set a step limit (not shown in the algorithm) since bad actions can cause the solver to fail to converge.
During training, we also always adapt $\rho$ in each step and ignore the heuristic adapt/no-adapt policy.

For well-scaled QPs, the residuals and $\rho$ can reasonably range between $10^{-6}$ and $10^{6}$.  Since this can cause issues with training the policy networks, we train the policy network with logs of the residuals, and exponentiate the network's output to get the action to apply.

\subsection{RL Policy for Vector Coefficient Adaptation}

For some classes of QPs, the solver can further speed up convergence by adapting all coefficients of of the vector $\rho$, instead of applying Eq.~\ref{eqn:rhobartorho} to a scalar $\bar\rho$.  Conceptually, this could be accomplished with a policy $\pi_\mathrm{vec} : S_\mathrm{QP} \rightarrow A_\mathrm{vec}$, where $S_\mathrm{QP} \in \mathbb{R}^{O(n+m)}$ is the internal state of the solver and $A_\mathrm{vec} \in \mathbb{R}_+^m$ is the new value for $\rho$.  However, due to variation in problem size and permutation, we instead propose a simplification in which $\pi_\mathrm{vec}$ is formulated as a policy $\pi_\rho : S_\rho \rightarrow A_\rho$ that is applied per coefficient of $\rho$.  Here, $S_\rho \in \mathbb{R}^6$ is state corresponding to a single coefficient in $\rho$, and $A_\rho \in \mathbb{R}$ is the value to set for that coefficient.

To define $S_\rho$, we observe that coefficients in $\rho$ are one-to-one with coefficients in ${y}$, ${z}$, ${l}$, ${u}$, and $Ax$.
We observe that constraint bounds are likely to have an impact on an ADMM iteration when coefficients of ${z}$ are ``close'' to their bounds in ${l}$ or ${u}$.
A coefficient in ${z}$ is also ``close'' to a solution when it is nearly equal to the corresponding coefficient in ${Ax}$.  Finally, to include a permutation-invariant signal on the overall convergence,
we include the primal and dual residuals of the QP solver; these are infinity norms of individual residuals, and is similar to using a max-pooling operation on the input to a graph neural network~\cite{scarselli2008graph,battaglia2018relational}.
We thus define a coefficient's state as:
\[
s_i = 
\begin{bmatrix}
\min (z_i - l_i, u_i - z_i) \\
z_i - (Ax)_i \\
y_i \\
\rho_i \\
\xi_\mathrm{primal} \\
\xi_\mathrm{dual}
\end{bmatrix}
\in S_\rho.
\]
In practice, %
we clamp values in each state $s_i$ to reasonable ranges (e.g., $[10^{-8}, 10^6]$, $[-10^6, 10^6]$, $[-10^6, 10^6]$, $[10^{-6},10^6]$, $[10^{-6},10^6]$, $[10^{-6},10^6]$ for the coefficients of $s_i$, in order). Empirically, training is more efficient if the policy operates on states with the log of the first and last 3 coefficients. %

Since each $\mathtt{step}$ in the vector formulation applies $m$ actions and updates $m$ states simultaneously, we adapt the multi-agent single-policy TD3 formulation from \citet{huang2020one}, and show it in Alg.~\ref{alg:ddpg_vec}, with the main differences from Alg.~\ref{alg:ddpg} highlighted in blue.
Before each step, $\mathtt{step}$ applies the policy with exploration noise to generate $m$ actions (coefficient updates to $\rho$).
After each $\mathtt{step}$, Alg.~\ref{alg:ddpg_vec} adds the $m$ states before the action, the $m$ actions, and the $m$ states after the action, along with the single reward to the replay buffer.  Since each step results in $m$ tuples added to the replay buffer, Alg.~\ref{alg:ddpg_vec} allocates a replay buffer large enough to hold the average number of tuples that each QP in the training set will have.

The hypothesis of this approach is that the some coefficients, and thus policy actions for coefficients, will have more of an effect on convergence, and thus the reward, than others.  When the domain for the policy function has more of an effect, the range of the actions will have lower variance.  Similarly, when the policy values has less effect, the variance will be higher.  This suggests that when training the policy network in this case, having a lower learning rate, and higher batch size can help.  A lower learning rate will cause smaller gradient steps when training the network so that it does not overfit to some part of the high variance training data.  A higher batch size will allow gradients to average out in high variance training data so that the gradient step better matches the true mean of the data. %

\section{Experiments}
\label{sec:experiments}

To train and test the proposed methods, we modify OSQP
to support direct querying and modification of its $\rho$ vector, and integrate both $\pi_{\bar\rho}$ and $\pi_\mathrm{vec}$ policies for benchmarking, and a runtime flag to switch between policies.
We train the network using randomly generated QPs from OSQP's benchmark suite.  The form of these QPs falls into 7 classes (see below), but the specific coefficient values in the objective and constraints are generated from a random-number generator.
These QPs are also guaranteed to be feasible by construction (e.g., by reverse engineering constraint values from a pre-generated solution).
To separate train and test sets, we ensure that each set is generated from uniquely seeded random-number generators.
Training is performed in PyTorch with a python wrapper around the modified OSQP which is written C/C++.
During benchmarking, the solver performs runtime adaptation of $\rho$ using PyTorch's C++ API on the already-trained policy network.
We train a small model to keep runtime network inference as fast as possible.

We evaluate all policies with 7 problem domains (referred to as the ``benchmark problem'') defined in Appendix A of the paper on OSQP~\cite{osqp}.  These policies cover control, Huber fitting, support-vector machines (SVM), Lasso regression, Portfolio optimization, equality constrained, and random QP domains. Alongside \ours{}, we benchmark the unmodified OSQP solver to evaluate how the RL policy improves convergence.  While our focus is on improving the first-order method in OSQP with an RL policy, we include some benchmarks against the state-of-the-art commercial Gurobi solver~\cite{Gurobi} as it may be of interest to a practitioner.  %

We consider three evaluation configurations: (1) \emph{multi-task policy} learning in which we train a single \ours{} policy on a suite of random benchmark problems and test it across all problems, (2) \emph{class-specific policy} learning in which we train and test the policy for a single problem domain and (3) \emph{zero-shot generalization} where we test a general policy on a novel unseen problem class.

We evaluate speedups with the shifted geometric mean~\cite{gould2016note_shiftedgeomean} as problems have wide variations in runtime across several orders of magnitude. This metric is the standard benchmark used by optimization community. The shifted geometric mean is computed as:
\[ %
\exp {\sum_{i=1}^N (1/N) \log(\max(1, v_i + s))} - s,
\] %
where $v_i$ is compute time in seconds, $s = 10$, and $N$ is the number of values (e.g., QPs solved).%

We also evaluate on QPLIB~\cite{FuriniEtAl2018online}, Netlib~\cite{netlib_1985}, and Maros and M{\'e}sz{\'a}ros~\cite{maros1999repository}, as they are well-established benchmark problems in the optimization community.

In all experiments, the policy network architecture has 3 fully-connected hidden layers of 48 with ReLU activations between the input and output layers.  The input layer is normalized, and the output activation is Tanh. The critic network architectures uses the identity function as the output activation, but otherwise matches the policy.  As small networks for fast CPU inferences are desirable here, we attempted to keep the network as small as possible.  We performed minimal experimentation before settling on this architecture---finding that smaller networks fail to converge during training.  

We trained on a system with 256\,GiB RAM, two Intel Xeon E5-2650 v4 CPUs @ 2.20\,GHz for a total of 24 cores (48 hyperthreads), and five NVIDIA Tesla V100s. We ran benchmarks on a system with Intel i9 8-core CPU @ 2.4~GHz and without GPU acceleration.

\subsection{Multi-task/General \ours{} Policy}
\begin{figure}
    \centering
    \begin{tabular}{@{}c@{}c@{}}
    \includegraphics[width=3in]{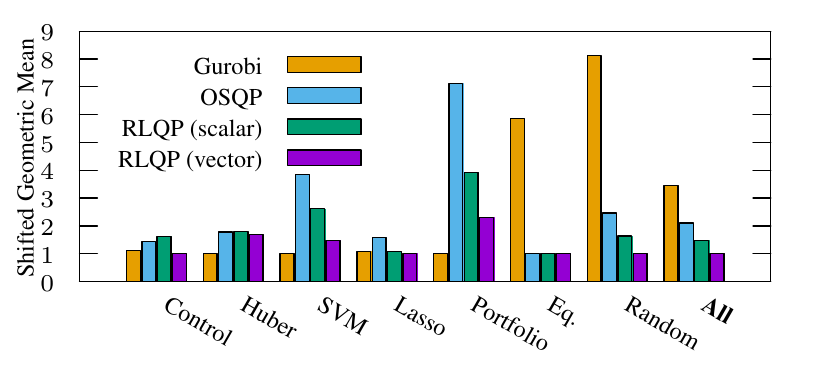}&%
    \includegraphics[width=2.5in]{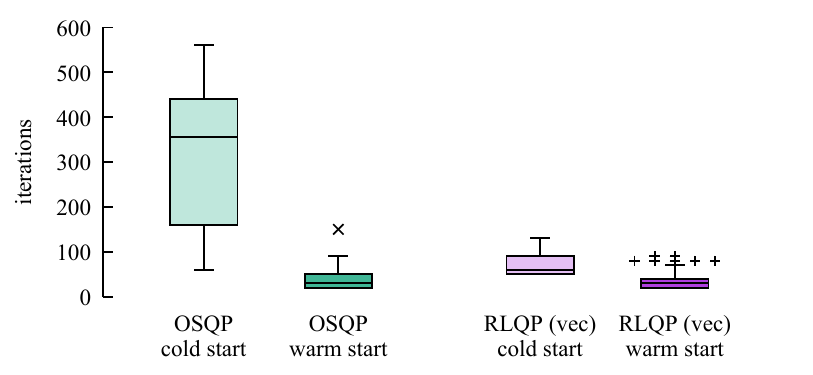}%
    \end{tabular}
    \caption{\textbf{Left:} Comparison of \emph{general} adaptation policy applied to different classes.  We train an RL policy using multiple classes, and show the performance per class, along with each class.  The $y$-axis is the shifted geometric mean across problems within each class, and the value of 1 is always assigned to the best in class.  The right-most \textbf{All} class is the aggregate of all classes to the left of it.
    \textbf{Right:} Comparison of warm-starting performance using OSQP's warm-start benchmark.}
    \label{fig:generic_policy_class_results}
\end{figure}

We train a general policy on a broad set of problem classes and compare solve times with different classes. During training, we sample one of seven QP domains from benchmark problem. From that sampled problem domain, we generate a random problem.

In Fig.~\ref{fig:generic_policy_class_results}, we 
compare the shifted geometric mean of solving 10 problems of 20 different dimension, for a total of 200 runs per class per solver.  The problem dimensions for Control, Huber, SVM, Lasso are (10,  11,  12,  13,  14,  16,  17,  20,  23,  26,  31,  37,  45,
        55,  68,  84, 105, 132, 166, 209);
for Random and Eq are (10,   11,   12,   13,   15,   18,   23,   29,   39,   53,   73,
        103,  146,  211,  304,  442,  644,  940, 1373, 2009), and for Portfolio are (5,   6,   7,   8,   9,  10,  12,  14,  16,  20,  24,  28,  35,
        43,  52,  65,  80,  99, 124, 154).
From the results, we observe that both RLQP adaptation policies typically improve upon convergence rate from the handcrafted policy in OSQP, and in some cases, e.g. Portfolio optimization, by up to 3x.

\subsection{Problem Dimension Scaling}
\begin{figure}
    \centering
    \includegraphics[width=0.8\linewidth]{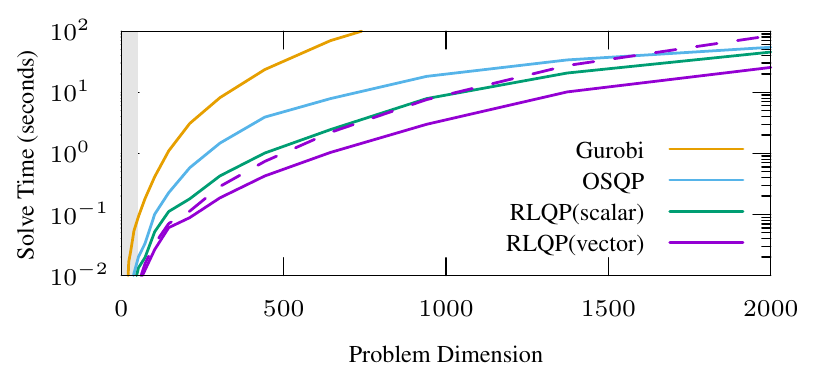}
    \caption{Solve time with increasing dimension on the Random QP problem set.
    We train and benchmark two vector RL adaptation policies: (dashed) on problems ranging from dimension 10 to 50, and (solid) on problems ranging from 10 to 2000.  The gray box shows the range of the training data for the dashed line.  When the benchmark run is in the same problem-dimension distribution as the training data, the relative performance between solvers is consistent, however, when the problem dimension is outside of he training distribution, the performance diverges.}
    \label{fig:dimension_scaling_results}
\end{figure}

To test how a trained policy scales to higher dimensions, we train a policy on low dimensional problems (10 to 50), and solve problems with varying dimensions, including dimensions higher than the training set (up to 2000).  For comparison, we also include a policy trained on the full dimension range (10 to 2000).  From the results plotted in Fig.~\ref{fig:dimension_scaling_results}, we observe that a policy trained on a lower dimensional training set, can show improvement beyond its training range.  However, as the problem size diverges more from the training set, its performance suffers and it eventually loses to the handcrafted policy.  Both low-dimensional and full-dimension range polices, were trained using the same network architecture, we hypothesize that this behavior is a function of the training data and not a limitation of the network expressiveness.  While this is a disadvantage of using smaller problems for training, in practice it may be outweighed by the advantage in training time---as each RL step requires $O((n+m)^3)$ compute time.

\subsection{Training a Class-Specific Policy}

Many applications in control~\cite{ichnowski2020gomp} and optimization~\cite{mlsys2020_196} require QPs from the same class to be repeatedly solved.  To test if training a policy specific to a QP class can outperform a policy trained on the benchmark suite, we train policies specific to the 
problems generated by the trust-region~\cite{conn2000trust} based solver for sequential quadratic program (SQP) from a grasp-optimized motion planner (GOMP)~\cite{ichnowski2020gomp,ichnowski2020djgomp} for robots.
With these problems, RLQP trained on the benchmarks converges more slowly than the handcrafted policy included in OSQP.  With a vector policy trained on the the QPs from the SQP, the shifted geometric mean of OSQP is 1.37.  This result suggests that while a general policy may work for multiple problem classes, there are cases in which it is beneficial to train a policy specific to a problem class, particularly if the QPs from that problem class are repeatedly solved.

\subsection{Warm Starting QPs}
One benefit of first-order method such as OSQP is their ability to warm start---that is, rapidly converge from a good initial guess.  We test if RLQP retains the benefit of warm start on OSQP's warm-start benchmark and show the results in Fig.~\ref{fig:generic_policy_class_results} (right).
As warm starts require fewer iterations, and thus fewer adaptations than cold starts, we expect RLQP to show a smaller improvement here.
In the plot, we can see that RLQP retains the benefit of warm starting, and also gains a improvement over OSQP.

\balance
\subsection{QPLIB}
We benchmark convex continuous QP instances with constraints from QPLIB~\cite{FuriniEtAl2018online}, and show the results in Table~\ref{tab:qplib}.  Since there are only a few such QPLIB instances and they come from varying classes, creating a train/test split is problematic.  We thus use the general policy trained on the benchmark classes.
From the table, we observe that the general RLQP policy beats OSQP's heuristic policy in all but three cases.  In two cases RLQP fails due to reaching an iteration or time limit. Training on similar problems should help avoid a timeout.
\begin{table}[]
    \centering
    \footnotesize
    \begin{tabular}{@{}c@{\quad}r@{\quad}r@{\quad}r rrr@{}}\toprule
                  &         &         &   non-  &           & RLQP     &  RLQP      \\ %
         Inst.    &     $n$ &     $m$ &   zeros &     OSQP  & (scalar) & (vector)   \\ %
         \midrule                                                                   
         \tt 8845 &    1546 &     777 &   10999 &     6.386 &   timeout & \bf  5.435 \\ %
         \tt 9002 &    2890 &    1649 &   12580 & \bf 6.000 &   timeout &     timeout \\ %
         \tt 8906 &    5223 &     838 &   20781 &     1.108 &    1.447 & \bf  0.741 \\ %
         \tt 8559 &   10000 &    5000 &   24998 &    59.648 &  205.372 & \bf 24.083 \\ %
         \tt 8938 &    4001 &   11999 &   31997 &    timeout &   timeout & \bf  0.991 \\ %
         \tt 8567 &   10000 &    7500 &   32497 &    98.511 &  284.112 & \bf 22.222 \\ %
         \tt 8616 &   13870 &   10404 &   41610 &     0.126 & \bf0.113 &      0.141 \\ %
         \tt 8515 &   16002 &    8002 &.  56005 & \bf 0.105 &   timeout &     timeout \\ %
         \tt 8785 &   10399 &   11362 &   63023 &     6.334 &   timeout & \bf  2.972 \\ %
         \tt 8495 &   27543 &    8000 &   73029 &     1.612 &    0.742 & \bf  1.174 \\ %
         \tt 8602 &   34552 &   52983 &  242887 &    99.872 &   timeout & \bf 55.629 \\ %
         \tt 8547 & 1003001 & 1001000 & 6003001 &    timeout &   timeout &     timeout \\ %
         \bottomrule
    \end{tabular}
    \vspace{5.0mm}
    \caption{\textbf{QPLIB problems}. Timing results for solving the convex continuous QPs with constraints from QPLIB~\cite{FuriniEtAl2018online}. The Inst. column is QPLIB's instance number.  The columns $n$ (number of variables), $m$ (number of constraints), and non-zeros indicate the QP's complexity. A \emph{timeout} result indicates the solver terminated due to reaching an iteration or time limit (300~s).  We hypothesize that the RLQP timeouts are due to out of distribution test problems, as the policy here was trained on the benchmark classes.}
    \label{tab:qplib}
\end{table}

\subsection{Netlib Linear Programming benchmark}
The Netlib Linear Programming benchmark~\cite{netlib_1985} contains 98 challenging real-world problems including supply-chain optimization, scheduling and control problems. As with the QPLIB benchmark, we evaluate results with a general policy trained on the benchmark classes. We solve problems to high-accuracy as many of these benchmarks are poorly scaled. Overall, vector formulation of \ours{} is 1.30$\times$ faster than OSQP by the scaled geomean of runtimes. We include a problem-specific breakdown in the supplementary materials.

\subsection{Maros and M{\'e}sz{\'a}ros}
In a manner similar to the QPLIB problems, we also benchmark on the Maros and M{\'e}sz{\'a}ros repository of QPs.~\cite{maros1999repository}.
This collection of 138 QP problems, includes many poorly scaled problems that cause OSQP to fail to converge.  We compute the shifted geometric mean for problems solved by both OSQP and RLQP with the general vector policy.  RLQP converges faster, with OSQP's shifted geometric mean is 1.829 times that of RLQP.  Because the dataset contains 138 problems, a table of the full results is included in the Supplementary Material.
\section{Limitations}
\label{sec:limitations}

\ours{} has limitations.
For QPs that converge after few iterations, and thus do not adapt $\rho$, having a better adaptation policy is moot.
Training RLQP can take a prohibitively long time and require a large replay buffer for some applications, for example, to train the benchmark suite of QPs required several days on a high-end computer with 256\,GiB---this may be mitigated to an extent by sharing learned policies between interested practitioners.
The time it takes to evaluate the RL policy, especially the vector version, may reduce the performance benefit of faster convergence---this may be mitigated by learning more efficient representations, or by using dedicated neural-network processing hardware.

\section{Conclusion}
\label{sec:conclusion}

We presented RLQP, a method for using reinforcement learning (RL) to speed up the convergence rate of a first-order method quadratic program solver.
RLQP uses RL to learn a policy to adapt the internal parameters of the solver to allow for fewer iterations and faster convergence.
In experiments, we trained a generic policy and results suggest that a single policy can improve convergence rates for a broad class of problems.
Results for a problem-specific policy suggest that fine-tuning can further accelerate convergence rates.

In future work, we will explore whether additional RL policy options can speed up convergence rate further, such as training a hierarchical policy~\cite{barto2003recent} in which the higher-level policy determines the interval between adaptation, performing a neural-architecture search~\cite{elsken2019neural}, using meta-learning~\cite{finn2017model,nichol2018first} to speed up problem-specific training, and online-learning to adjust the policy at runtime to adapt to changing problems.

\section*{Acknowledgements}
This research was performed at the AUTOLAB at UC Berkeley in affiliation with the Berkeley AI Research (BAIR) Lab, and the CITRIS ``People and Robots" (CPAR) Initiative. 
In addition to NSF CISE Expeditions Award CCF-1730628, this research is supported by gifts from Amazon Web Services, Ant Group, Ericsson, Facebook, Futurewei, Google, Intel, Microsoft, Nvidia, Scotiabank, Splunk and VMware.
Any opinions, findings, and conclusions or recommendations expressed in this material are those of the authors and do not necessarily reflect the views of the sponsors.
We thank our colleagues who provided helpful feedback and suggestions, in particular Ashwin Balakrishna and Arnav Gulati.

\bibliographystyle{plainnat}
\bibliography{main}

\clearpage
\appendix
\section{Implementation}

Training the scalar policy for OSQP~\cite{osqp} requires no modification of the OSQP source code.  Instead, we disable the builtin {\tt adaptive\_rho} setting and set {\tt max\_iter} and {\tt check\_termination} to the interval to associate with the policy (e.g., 100).  With these settings, the solver will run for the preset iteration count and either return ``solved'' or ``iteration limit reached.''  Upon reaching the iteration limit, the RL policy step applies the adaptation via an existing call.  On the subsequent step, the internal state of the QP solver remains otherwise unchanged, thus this process mimics adapting the $\rho$ in the inner loop of he solver.

Training the vector policy requires a minor modification of OSQP to support setting and getting the internal $\rho$ vector.  Otherwise, training the vector policy is the same as training the scalar policy.

Using and benchmarking the policy requires additional modification of the solver. We modify the code so that when the {\tt adaptive\_rho} setting is enabled, OSQP calls through the PyTorch C++ API~\cite{pytorch} to pass the internal state through the learned policy network and then apply the adaptation internally.

We parallelize the training implementation to run multiple episodes concurrently, but otherwise follow close to the TD3~\cite{fujimoto2018addressing} algorithm for the scalar policy, and according to the one-policy~\cite{huang2020one} modifications described in the main text.  %
When training reaches an update or epoch step, the implementation waits for concurrently running episodes to complete before updating the networks---this leads to imprecise step counts between training, but does not appear to otherwise effect training.

\begin{figure}[t!]
    \centering
    \includegraphics[width=3.25in]{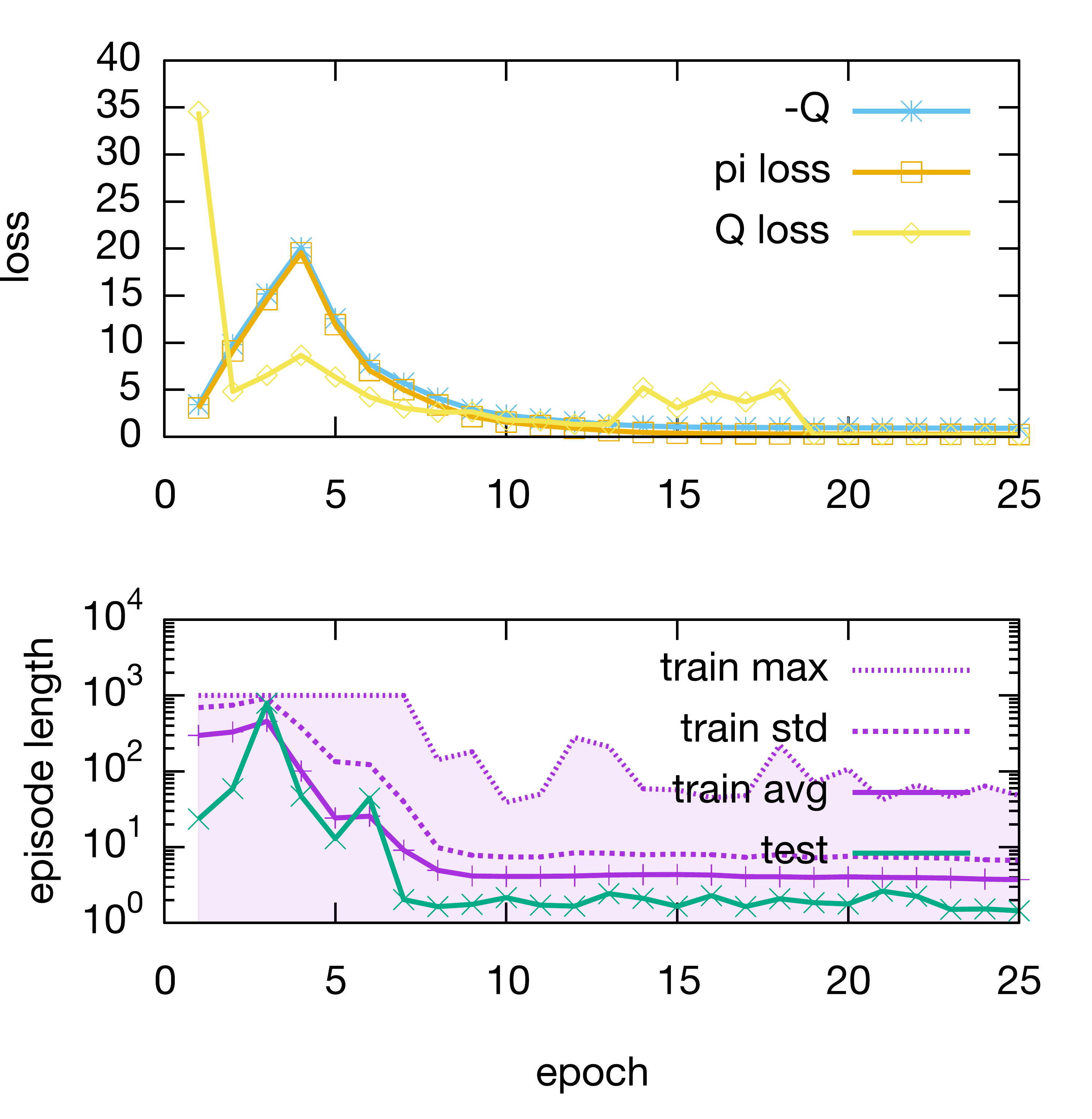}
    \caption{{\bf Reinforcement learning training curves.}  In these plots, we show the training curves over a training run.  The top graph shows the policy (pi) and critic (Q) loss, along with the negated average critic (-Q) value.  The bottom graph shows the training episode length maximum (train max), average length + standard deviation (train std), and average length (train avg), and the test episode average.  The top graph converges to smaller loss indicating that the policy and critic are improving.  The bottom graph shows that average and maximum episode length lowers as training continues.}
    \label{fig:rlqp_training_curves}
\end{figure}

We plot the training curves on learning the benchmark problems in Fig.~\ref{fig:rlqp_training_curves}.  In this figure we observe that the policy and critic loss lowers over training time, and correspondingly that the episode length (which is the negative reward), goes down as the learned policy improves.

\section{Comparison and Ablation of Training and Policies}

\begin{figure}[t]
    \centering
    \includegraphics{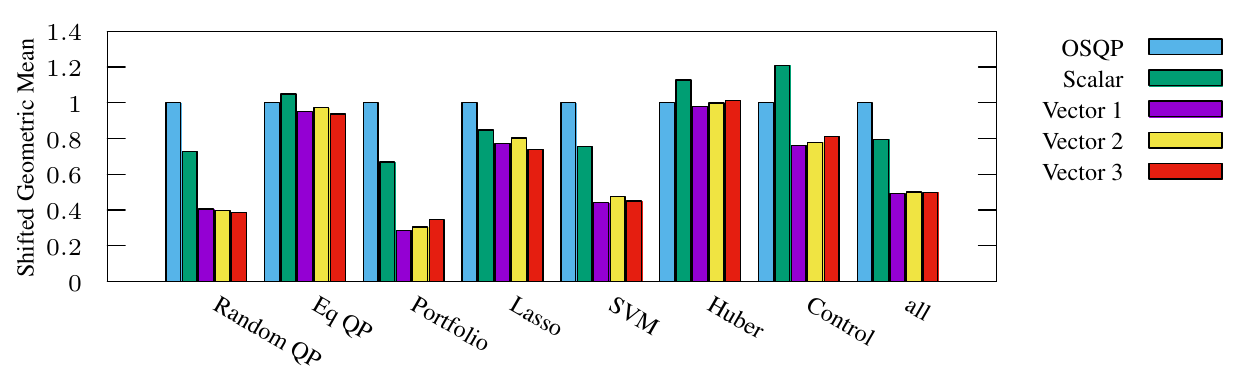}
    \caption{Comparison of the geometric mean of solve times for policies from different training runs.  Here we normalize to the geometric mean of OSQP at 1.0.  See text for description of the policies and how they were trained.}
    \label{fig:models}
\end{figure}

We compare multiple training runs with different seeds for different model architectures, and plot the results in Fig.~\ref{fig:models}.
The \emph{Vector 1} policy does not include residuals $\xi_\mathrm{primal}$ and $\xi_\mathrm{dual}$ in $S$, while \emph{Vector 2} and \emph{Vector 3} policies do.  The \emph{Vector 1} and \emph{Vector 2} policies are networks with 3 hidden layers, while \emph{Vector 3} has 2 hidden layers, all layers are 48 wide with ReLU activations.  All policies were trained for a maximum of 50 epochs, with a replay buffer size of $4\times10^8$, $10^5$ initial steps, updates every 10000 steps, 5000 batch size, 20000 steps per epoch, 0.995 polyak, 1.0 noise, 2.5 noise clip, and policy updates every other critic update.  For 3-layer networks, we set the learning rate to $10^{-5}$ for both policy and critic networks, and for the 2-layer network, we set the learning rate to $10^{-6}$.  We selected the epoch with the lowest average loss, though better performance may be possible with a policy from a different epoch.  We observe minor variation in the 3 trained policies, but not sufficient to categorically state which one is the best.

\begin{figure}[t!]
    \centering
    \includegraphics[width=0.5\linewidth]{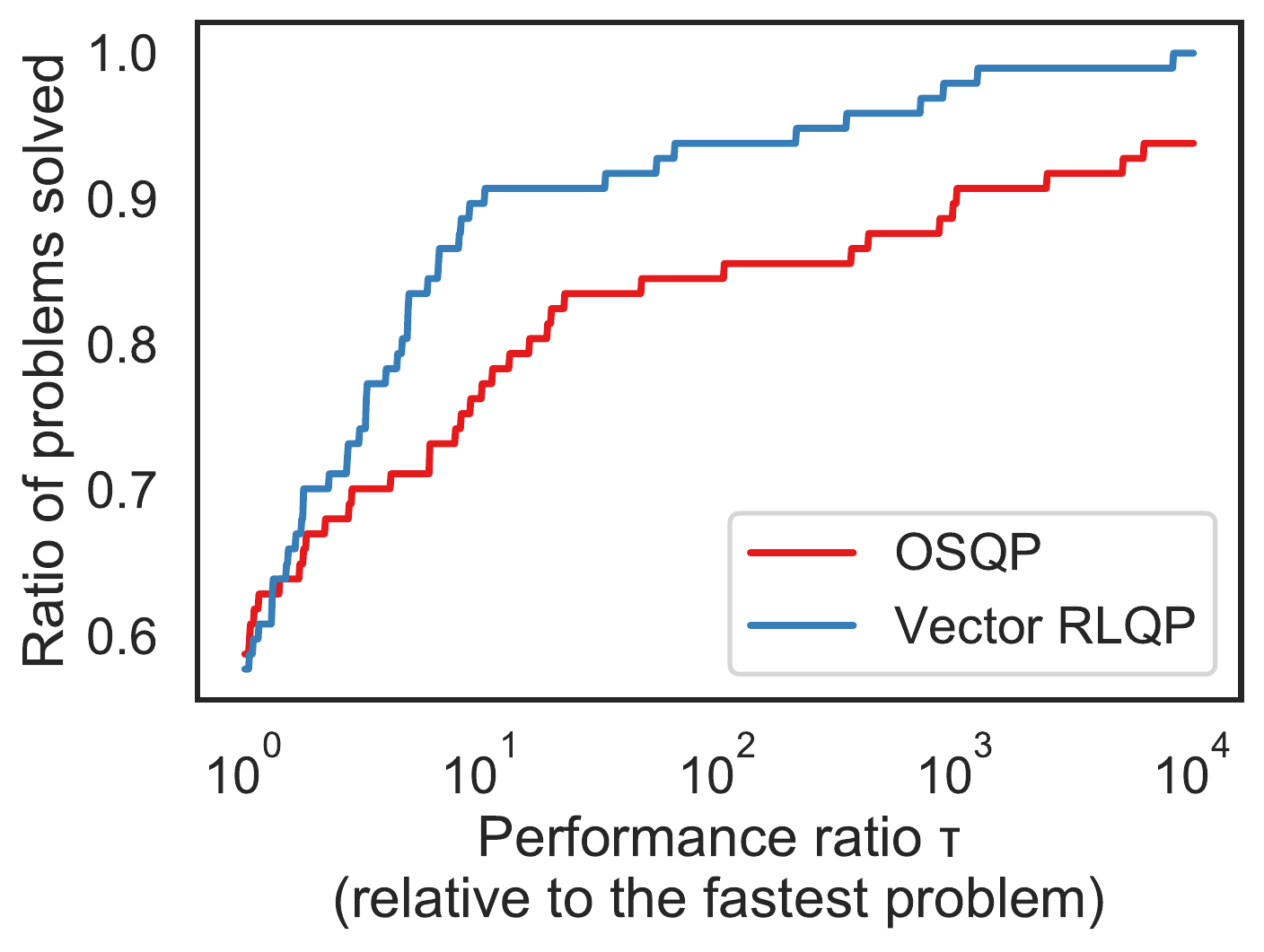}
    \caption{\textbf{Netlib LP performance profiles} We evaluate how the learned RLQP policy generalizes to unseen problems. The vector policy is $1.3\times$ faster (shifted geomean) than the existing heuristic in OSQP while solving 5.2\% more problems.}
    \label{fig:supplement_netlib_bench}
\end{figure}
\section{Netlib Linear Programming Results}
In order to measure how well the vector RL policy for OSQP generalizes to unseen inputs, we evaluate the policy on the 98 Netlib LP test problems~\cite{netlib_1985}. These problems are a collection of linear programs considered to be large and challenging. We select this benchmark as this class of linear programs is significantly different than any of the quadratic program classes we train with.

Overall, the vector RLQP policy outperforms the OSQP policy with a shifted geometric mean runtime that is $1.30\times$ faster. Moreover, the vector RLQP policy solves 5.2\% more problems than the heuristic OSQP. Figure~\ref{fig:supplement_netlib_bench} shows the number of problems solved by OSQP and RLQP with increasing runtime. Performance ratio ($\tau$) represents the rescaled runtime relative to the fastest problem, following the practice of~\citet{dolan2002benchmarking}.

These results are slightly better than the Netlib LP results included in the main paper. With the extra time, we were able to slightly tune the training procedure. Namely, we reduced the replay buffer size (which avoids training the policy with stale rollouts), decreased the learning rate, increased the batch size and finally trained the policy longer. These changes do not substantially change results (from $1.23\times$ to $1.30\times$). Moreover, the Netlib LP problems require a large number of iterations from the OSQP solver. We increased the maximum number of iterations for Netlib LP evaluation to $10^6$ iterations.

\begin{figure}[t]
    \centering
    \includegraphics[width=0.5\linewidth]{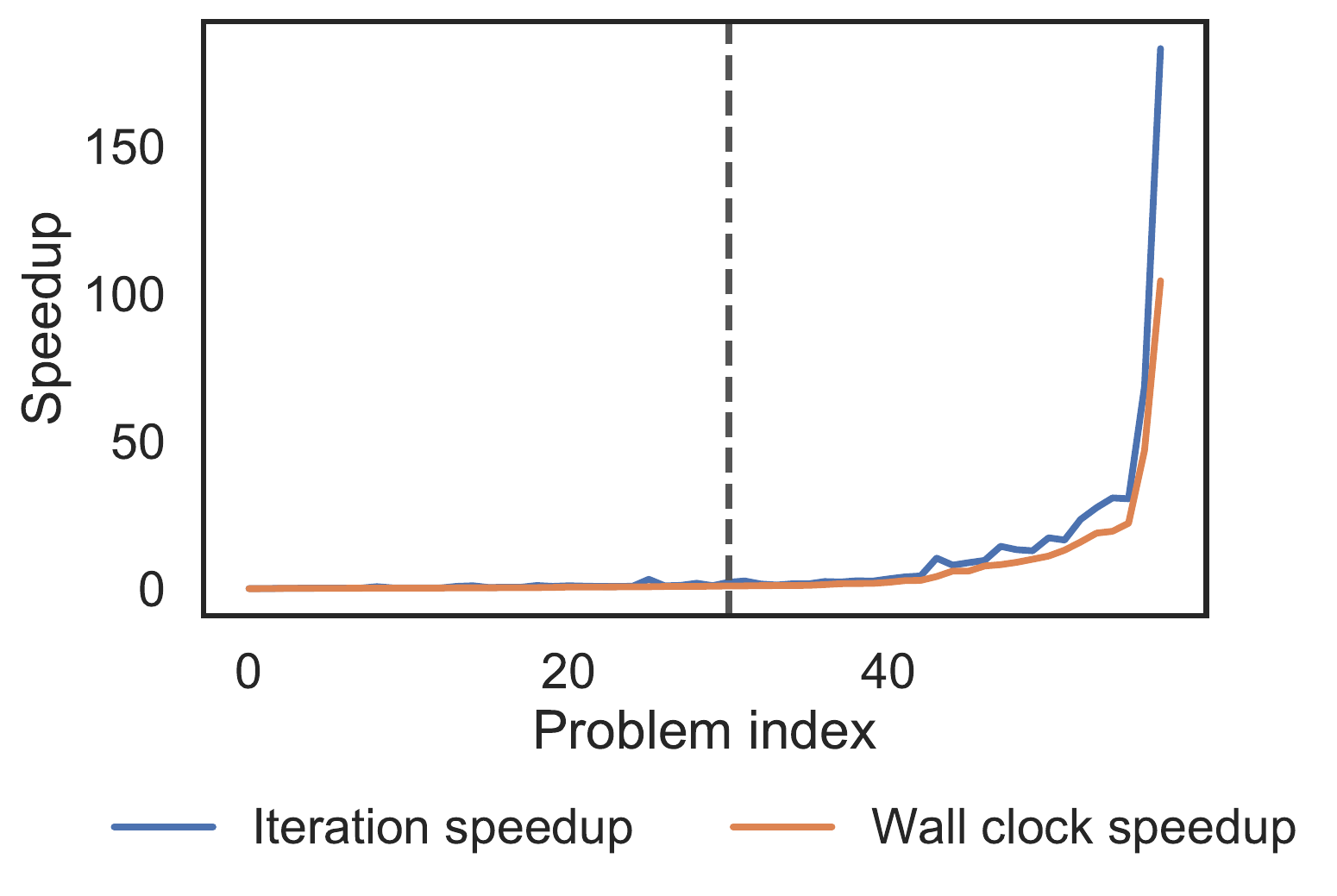}
    \caption{\textbf{Netlib LP problem speedup} Iteration speedup per problem in the Netlib LP problem set. Problems right of the dotted line observe speedup greater than 1. For the majority of problems, RLQP accelerates convergence by up to $73\times$.}
    \label{fig:supplement_netlib_speedups}
\end{figure}

While the vector RLQP policy accelerates Netlib LP optimization overall, it can slow convergence for some problems. In Figure~\ref{fig:supplement_netlib_speedups} displays per-problem speedups of RLQP over OSQP. RLQP achieves speedups of up to 73x, but degrades performance for a minority of problems. We include detailed per-problem results containing solver runtime in Section~\ref{tab:netlib_detailed}. As we evaluate the policy at fixed intervals, the solver must re-factorize the problem due to a change in $\rho$. However, the policy may update $\rho$ more times than is needed which can slow convergence for some fast well-conditioned problems. Our work is a good starting place for further research into learning methods for first-order optimization. We are extending the RLQP framework to support dynamic policy evaluation which would improve performance for these small-scale problems.

\section{Maros and M{\'e}sz{\'a}ros Results}

As with the Netlib linear problems, we evaluate the policy trained on the benchmark problems on all 138 Maros and M{\'e}sz{\'a}ros~\cite{maros1999repository} QP problems and present the results here.  We have made no effort to ensure that training problems come from the same distribution of QPs as the Maros and M{\'e}sz{\'a}ros problems.
Many of these QPs are poorly scaled, which causes both OSQP and RLQP to sometimes fail to converge within a 600\,s time limit we set.  Some problems that OSQP fails to solve, RLQP (vector) solves, and vice versa, while the (scalar) policy performs poorly on most of these problems (not shown).  We show results for two (vector) models trained on the benchmarks.  The ``GNN'' model includes the primal and dual residuals ($\xi_\mathrm{primal}$ and $\xi_\mathrm{dual}$) in $S$, while the ``non-GNN'' does not.
In the table that follows, the bold entries are the fasted solve times in seconds and the fewest ADMM iterations, though we omit the bold when the three policies tie.  We report the number of times OSQP and RLQP have the fastest solve time and fewest iterations, and observe that the difference between these indicates that time to compute the adaptation is a factor in making RLQP not outperform OSQP more often.

\clearpage
\section{Detailed results for Netlib LP problems}
\label{tab:netlib_detailed}

\tablehead{%
\toprule
Netlib LP         &         &         &   non-  &            & RLQP \\
Problem           &     $n$ &     $m$ &   zeros &     OSQP   & (vector) \\
\midrule
}
\tabletail{%
\midrule
\multicolumn{6}{r}{\scriptsize continued \ldots} \\
\bottomrule}

\tablelasttail{\bottomrule}

{
\centering
\scriptsize
    \begin{supertabular}{@{}l@{\quad}r@{\quad}r@{\quad}r rr@{}}%

  \texttt{25FV47} &  1876 &  2697 &       12581 &   \textbf{3.496} &         31.064 \\
 \texttt{80BAU3B} & 12061 & 14323 &       35325 &  \textbf{11.569} &         52.989 \\
\texttt{ADLITTLE} &   138 &   194 &         562 &   \textbf{0.076} &          0.079 \\
   \texttt{AFIRO} &    51 &    78 &         153 &   \textbf{0.001} &          0.002 \\
    \texttt{AGG2} &   758 &  1274 &        5498 &           timeout & \textbf{1.183} \\
    \texttt{AGG3} &   758 &  1274 &        5514 &           timeout & \textbf{0.415} \\
     \texttt{AGG} &   615 &  1103 &        3477 &           timeout &         timeout \\
   \texttt{BANDM} &   472 &   777 &        2966 &            0.466 & \textbf{0.264} \\
\texttt{BEACONFD} &   295 &   468 &        3703 &            0.025 & \textbf{0.024} \\
   \texttt{BLEND} &   114 &   188 &         636 &            0.031 & \textbf{0.007} \\
    \texttt{BNL1} &  1586 &  2229 &        7118 &           timeout & \textbf{0.998} \\
    \texttt{BNL2} &  4486 &  6810 &       19482 &  \textbf{24.329} &         37.051 \\
 \texttt{BOEING1} &   726 &  1077 &        4553 &            3.119 & \textbf{0.348} \\
 \texttt{BOEING2} &   305 &   471 &        1663 &           timeout & \textbf{0.198} \\
  \texttt{BORE3D} &   334 &   567 &        1782 &            0.585 & \textbf{0.419} \\
  \texttt{BRANDY} &   303 &   523 &        2505 &   \textbf{0.548} &          0.962 \\
   \texttt{CAPRI} &   496 &   767 &        2461 &            4.846 & \textbf{0.437} \\
   \texttt{CYCLE} &  3378 &  5281 &       24626 &   \textbf{4.931} &         29.043 \\
  \texttt{CZPROB} &  3562 &  4491 &       14270 &           10.714 & \textbf{1.388} \\
  \texttt{D2Q06C} &  5831 &  8002 &       38912 & \textbf{127.159} &        167.348 \\
  \texttt{D6CUBE} &  6184 &  6599 &       43888 &            3.211 & \textbf{0.321} \\
  \texttt{DEGEN2} &   757 &  1201 &        4958 &   \textbf{0.089} &          0.583 \\
  \texttt{DEGEN3} &  2604 &  4107 &       28036 &   \textbf{0.730} &          3.558 \\
  \texttt{DFL001} & 12230 & 18301 &       47862 &  \textbf{14.112} &        765.502 \\
    \texttt{E226} &   472 &   695 &        3240 &   \textbf{0.371} &          1.126 \\
\texttt{ETAMACRO} &   816 &  1216 &        3353 &   \textbf{0.655} &          6.718 \\
\texttt{FFFFF800} &  1028 &  1552 &        7429 &           timeout &         timeout \\
  \texttt{FINNIS} &  1064 &  1561 &        3824 &   \textbf{2.034} &          2.657 \\
   \texttt{FIT1D} &  1049 &  1073 &       14476 &   \textbf{0.390} &          1.895 \\
   \texttt{FIT1P} &  1677 &  2304 &       11545 &            0.478 & \textbf{0.080} \\
   \texttt{FIT2D} & 10524 & 10549 &      139566 &   \textbf{3.622} &        119.416 \\
   \texttt{FIT2P} & 13525 & 16525 &       63809 &   \textbf{0.533} &          2.332 \\
 \texttt{FORPLAN} &   492 &   653 &        5126 &            0.061 & \textbf{0.053} \\
  \texttt{GANGES} &  1706 &  3015 &        8643 &   \textbf{4.741} &         timeout \\
\texttt{GFRD-PNC} &  1160 &  1776 &        3605 &            0.790 & \textbf{0.288} \\
\texttt{GREENBEA} &  5598 &  7990 &       36668 &           timeout &         timeout \\
\texttt{GREENBEB} &  5602 &  7994 &       36677 &          122.834 &         timeout \\
  \texttt{GROW15} &   645 &   945 &        6265 &           timeout &         timeout \\
  \texttt{GROW22} &   946 &  1386 &        9198 &   \textbf{1.132} &         timeout \\
   \texttt{GROW7} &   301 &   441 &        2913 &           timeout &         timeout \\
  \texttt{ISRAEL} &   316 &   490 &        2759 &           timeout & \textbf{2.781} \\
     \texttt{KB2} &    68 &   111 &         381 &           timeout & \textbf{0.066} \\
   \texttt{LOTFI} &   366 &   519 &        1502 &            1.599 & \textbf{0.196} \\
\texttt{MAROS-R7} &  9408 & 12544 &      154256 & \textbf{253.193} &         timeout \\
   \texttt{MAROS} &  1966 &  2812 &       12103 &           timeout &         timeout \\
 \texttt{MODSZK1} &  1622 &  2309 &        4792 &   \textbf{1.588} &          5.152 \\
    \texttt{NESM} &  3105 &  3767 &       16575 &   \textbf{0.811} &         timeout \\
  \texttt{PEROLD} &  1594 &  2219 &        8911 &           timeout &         timeout \\
\texttt{PILOT-JA} &  2355 &  3295 &       18571 &           timeout &         timeout \\
\texttt{PILOT-WE} &  3008 &  3730 &       12809 &           timeout &         timeout \\
  \texttt{PILOT4} &  1211 &  1621 &        8553 &           timeout &         timeout \\
 \texttt{PILOT87} &  6680 &  8710 &       81629 &           timeout &         timeout \\
\texttt{PILOTNOV} &  2446 &  3421 &       15777 &           timeout &         timeout \\
   \texttt{PILOT} &  4860 &  6301 &       49235 &           timeout &         timeout \\
   \texttt{QAP12} &  8856 & 12048 &       47160 &   \textbf{9.819} &         26.535 \\
   \texttt{QAP15} & 22275 & 28605 &      117225 &  \textbf{91.608} &        137.196 \\
    \texttt{QAP8} &  1632 &  2544 &        8928 &            0.386 & \textbf{0.177} \\
\texttt{RECIPELP} &   204 &   295 &         891 &   \textbf{0.002} &          0.003 \\
   \texttt{SC105} &   163 &   268 &         503 &   \textbf{0.011} &          0.014 \\
   \texttt{SC205} &   317 &   522 &         982 &           timeout & \textbf{0.022} \\
   \texttt{SC50A} &    78 &   128 &         238 &   \textbf{0.003} &          0.009 \\
   \texttt{SC50B} &    78 &   128 &         226 &   \textbf{0.005} &          0.023 \\
 \texttt{SCAGR25} &   671 &  1142 &        2396 &   \textbf{0.122} &         timeout \\
  \texttt{SCAGR7} &   185 &   314 &         650 &   \textbf{0.081} &          0.087 \\
  \texttt{SCFXM1} &   600 &   930 &        3332 &   \textbf{2.895} &         timeout \\
  \texttt{SCFXM2} &  1200 &  1860 &        6669 &           timeout &         timeout \\
  \texttt{SCFXM3} &  1800 &  2790 &       10006 &  \textbf{15.458} &         timeout \\
\texttt{SCORPION} &   466 &   854 &        2000 &           timeout &         timeout \\
   \texttt{SCRS8} &  1275 &  1765 &        4563 &   \textbf{1.156} &          7.543 \\
   \texttt{SCSD1} &   760 &   837 &        3148 &            0.021 & \textbf{0.008} \\
   \texttt{SCSD6} &  1350 &  1497 &        5666 &            0.262 & \textbf{0.017} \\
   \texttt{SCSD8} &  2750 &  3147 &       11334 &            0.187 & \textbf{0.031} \\
  \texttt{SCTAP1} &   660 &   960 &        2532 &            1.492 & \textbf{0.014} \\
  \texttt{SCTAP2} &  2500 &  3590 &        9834 &            1.094 & \textbf{0.056} \\
  \texttt{SCTAP3} &  3340 &  4820 &       13074 &            1.192 & \textbf{0.054} \\
    \texttt{SEBA} &  1036 &  1551 &        5396 &            1.022 & \textbf{0.939} \\
 \texttt{SHARE1B} &   253 &   370 &        1432 &   \textbf{1.574} &          3.544 \\
 \texttt{SHARE2B} &   162 &   258 &         939 &           timeout & \textbf{0.030} \\
   \texttt{SHELL} &  1777 &  2313 &        5335 &            3.615 & \textbf{0.192} \\
 \texttt{SHIP04L} &  2166 &  2568 &        8546 &            0.716 & \textbf{0.397} \\
 \texttt{SHIP04S} &  1506 &  1908 &        5906 &   \textbf{0.091} &          0.730 \\
 \texttt{SHIP08L} &  4363 &  5141 &       17245 &   \textbf{0.372} &          0.608 \\
 \texttt{SHIP08S} &  2467 &  3245 &        9661 &           timeout & \textbf{1.034} \\
 \texttt{SHIP12L} &  5533 &  6684 &       21809 &            5.992 & \textbf{5.682} \\
 \texttt{SHIP12S} &  2869 &  4020 &       11153 &   \textbf{1.081} &          1.874 \\
  \texttt{SIERRA} &  2735 &  3962 &       10736 &            5.383 & \textbf{3.165} \\
   \texttt{STAIR} &   620 &   976 &        4641 &   \textbf{1.417} &         timeout \\
\texttt{STANDATA} &  1274 &  1633 &        4504 &           timeout & \textbf{0.075} \\
\texttt{STANDGUB} &  1383 &  1744 &        4722 &           timeout & \textbf{0.079} \\
\texttt{STANDMPS} &  1274 &  1741 &        5152 &            1.329 & \textbf{0.028} \\
\texttt{STOCFOR1} &   165 &   282 &         666 &           timeout & \textbf{0.013} \\
\texttt{STOCFOR2} &  3045 &  5202 &       12402 &   \textbf{2.599} &          7.081 \\
\texttt{STOCFOR3} & 23541 & 40216 &      100014 &           timeout &         timeout \\
   \texttt{TRUSS} &  8806 &  9806 &       36642 &           10.070 & \textbf{0.770} \\
\texttt{VTP-BASE} &   347 &   545 &        1399 &           timeout & \textbf{2.344} \\
  \texttt{WOOD1P} &  2595 &  2839 &       72811 &           timeout & \textbf{0.162} \\
   \texttt{WOODW} &  8418 &  9516 &       45905 &   \textbf{9.310} &         10.675 \\

\midrule
\multicolumn{4}{r}{\bf Total Solved:} & 67 & 72 \\
\end{supertabular} }

\clearpage
\section{Detailed results for Maros \& M{\'e}sz{\'a}ros problems}
\tablehead{%
\toprule
                                   &      &         &         & \multicolumn{3}{c}{Solve Time} & \multicolumn{3}{c}{Iteration} \\
        \cmidrule(lr){5-7}\cmidrule(l){8-10}
         Maros \& M{\'e}sz{\'a}ros &      &         &   non-  &           & RLQP     &  RLQP     & & RLQP & RLQP \\ %
         Problem    &     $n$ &     $m$ &   zeros &     OSQP  & non-GNN & GNN  & OSQP & non-GNN & GNN \\ %
         \midrule 
}
\tabletail{%
\midrule
\multicolumn{7}{r}{\scriptsize continued \ldots} \\
\bottomrule}

\tablelasttail{\bottomrule}

\bottomcaption{Detailed results for the Maros \& M{\'e}sz{\'a}ros problems~\cite{maros1999repository}.}

{
\centering
\scriptsize
    \begin{supertabular}{@{}l@{\quad}r@{\quad}r@{\quad}r rrr@{\qquad}rrr@{}}
\tt AUG2D & 20200 & 30200 & 80000 & \bf 0.155 & 0.164 & 0.163 & 200 & 200 & 200 \\
\tt AUG2DC & 20200 & 30200 & 80400 & \bf 0.153 & 0.188 & 0.155 & 200 & 200 & 200 \\
\tt AUG2DCQP & 20200 & 30200 & 80400 & 1.562 & 23.198 & \bf 0.939 & 2200 & 26800 & \bf 1000 \\
\tt AUG2DQP & 20200 & 30200 & 80000 & 1.683 & 8.923 & \bf 0.854 & 2400 & 10600 & \bf 1000 \\
\tt AUG3D & 3873 & 4873 & 13092 & \bf 0.028 & 0.039 & 0.037 & 200 & 200 & 200 \\
\tt AUG3DC & 3873 & 4873 & 14292 & \bf 0.026 & 0.031 & 0.035 & 200 & 200 & 200 \\
\tt AUG3DCQP & 3873 & 4873 & 14292 & \bf 0.056 & 0.063 & 0.065 & 400 & 400 & 400 \\
\tt AUG3DQP & 3873 & 4873 & 13092 & \bf 0.053 & 0.064 & 0.065 & 400 & 400 & 400 \\
\tt BOYD1 & 93261 & 93279 & 745507 & 286.552 & \bf 275.054 & timeout & 66000 & \bf 61400 & timeout \\
\tt BOYD2 & 93263 & 279794 & 517049 & timeout & timeout & timeout & timeout & timeout & timeout \\
\tt CONT-050 & 2597 & 4998 & 17199 & 0.395 & \bf 0.237 & 17.030 & 1600 & \bf 800 & 54800 \\
\tt CONT-100 & 10197 & 19998 & 69399 & 12.062 & \bf 1.766 & timeout & 8200 & \bf 1000 & timeout \\
\tt CONT-101 & 10197 & 20295 & 62496 & 20.508 & \bf 3.089 & timeout & 12800 & \bf 1800 & timeout \\
\tt CONT-200 & 40397 & 79998 & 278799 & 352.981 & \bf 87.121 & timeout & 33000 & \bf 7200 & timeout \\
\tt CONT-201 & 40397 & 80595 & 249996 & timeout & timeout & timeout & timeout & timeout & timeout \\
\tt CONT-300 & 90597 & 180895 & 562496 & timeout & timeout & timeout & timeout & timeout & timeout \\
\tt CVXQP1\_L & 10000 & 15000 & 94966 & 84.758 & \bf 31.133 & 104.432 & 9800 & \bf 1800 & 6200 \\
\tt CVXQP1\_M & 1000 & 1500 & 9466 & 0.161 & \bf 0.140 & 0.227 & 1200 & \bf 800 & 1400 \\
\tt CVXQP1\_S & 100 & 150 & 920 & \bf 0.004 & \bf 0.003 & 0.035 & 800 & \bf 600 & 6800 \\
\tt CVXQP2\_L & 10000 & 12500 & 87467 & 7.049 & 4.865 & \bf 4.748 & 800 & \bf 400 & \bf 400 \\
\tt CVXQP2\_M & 1000 & 1250 & 8717 & \bf 0.046 & 0.055 & 0.053 & 400 & 400 & 400 \\
\tt CVXQP2\_S & 100 & 125 & 846 & 0.001 & 0.001 & 0.001 & 200 & 200 & 200 \\
\tt CVXQP3\_L & 10000 & 17500 & 102465 & 99.156 & \bf 19.785 & 23.884 & 10200 & \bf 1000 & 1200 \\
\tt CVXQP3\_M & 1000 & 1750 & 10215 & 0.795 & \bf 0.444 & 40.058 & 5400 & \bf 2200 & 206400 \\
\tt CVXQP3\_S & 100 & 175 & 994 & \bf 0.002 & \bf 0.002 & 0.014 & \bf 400 & \bf 400 & 2200 \\
\tt DPKLO1 & 133 & 210 & 1785 & 0.002 & 0.002 & 0.003 & 200 & 200 & 200 \\
\tt DTOC3 & 14999 & 24997 & 64989 & 1.389 & \bf 0.191 & 7.221 & 3800 & \bf 400 & 16600 \\
\tt DUAL1 & 85 & 86 & 7201 & 0.002 & 0.002 & 0.002 & 200 & 200 & 200 \\
\tt DUAL2 & 96 & 97 & 9112 & 0.002 & 0.002 & 0.003 & 200 & 200 & 200 \\
\tt DUAL3 & 111 & 112 & 12327 & 0.003 & 0.003 & 0.004 & 200 & 200 & 200 \\
\tt DUAL4 & 75 & 76 & 5673 & 0.001 & 0.001 & 0.002 & 200 & 200 & 200 \\
\tt DUALC1 & 9 & 224 & 2025 & 0.002 & 0.002 & 0.002 & 600 & \bf 400 & \bf 400 \\
\tt DUALC2 & 7 & 236 & 1659 & 0.001 & 0.002 & 0.002 & 400 & 400 & 400 \\
\tt DUALC5 & 8 & 286 & 2296 & 0.001 & 0.001 & 0.001 & 200 & 200 & 200 \\
\tt DUALC8 & 8 & 511 & 4096 & \bf 0.002 & \bf 0.002 & 0.003 & 200 & 200 & 200 \\
\tt EXDATA & 3000 & 6001 & 2260500 & \bf 4.820 & 13.794 & 8.030 & \bf 2000 & 3200 & \bf 2000 \\
\tt GENHS28 & 10 & 18 & 62 & 0.000 & 0.000 & 0.000 & 200 & 200 & 200 \\
\tt GOULDQP2 & 699 & 1048 & 2791 & 0.020 & \bf 0.008 & 0.023 & 1400 & \bf 400 & 1200 \\
\tt GOULDQP3 & 699 & 1048 & 3838 & 0.003 & 0.004 & 0.004 & 200 & 200 & 200 \\
\tt HS118 & 15 & 32 & 69 & 0.000 & 0.000 & 0.000 & 800 & \bf 400 & \bf 400 \\
\tt HS21 & 2 & 3 & 6 & 0.000 & 0.000 & 0.000 & 200 & 200 & 200 \\
\tt HS268 & 5 & 10 & 55 & 0.000 & 0.000 & 0.000 & 400 & 400 & 400 \\
\tt HS35 & 3 & 4 & 13 & 0.000 & 0.000 & 0.000 & 200 & 200 & 200 \\
\tt HS35MOD & 3 & 4 & 13 & 0.000 & 0.000 & 0.000 & 200 & 200 & 200 \\
\tt HS51 & 5 & 8 & 21 & 0.000 & 0.000 & 0.000 & 200 & 200 & 200 \\
\tt HS52 & 5 & 8 & 21 & 0.000 & 0.000 & 0.000 & 200 & 200 & 200 \\
\tt HS53 & 5 & 8 & 21 & 0.000 & 0.000 & 0.000 & 200 & 200 & 200 \\
\tt HS76 & 4 & 7 & 22 & 0.000 & 0.000 & 0.000 & 200 & 200 & 200 \\
\tt HUES-MOD & 10000 & 10002 & 40000 & 0.223 & 0.174 & \bf 0.169 & 1200 & \bf 800 & \bf 800 \\
\tt HUESTIS & 10000 & 10002 & 40000 & 1.380 & \bf 0.269 & 54.088 & 7600 & \bf 1200 & 226600 \\
\tt KSIP & 20 & 1021 & 19938 & 0.058 & \bf 0.025 & 0.035 & 1800 & \bf 600 & 800 \\
\tt LASER & 1002 & 2002 & 9462 & \bf 0.011 & 0.012 & 0.014 & 400 & 400 & 400 \\
\tt LISWET1 & 10002 & 20002 & 50004 & 3.324 & 278.583 & \bf 0.851 & 11200 & 717600 & \bf 2400 \\
\tt LISWET10 & 10002 & 20002 & 50004 & 2.388 & 0.615 & \bf 0.312 & 8200 & 1600 & \bf 800 \\
\tt LISWET11 & 10002 & 20002 & 50004 & 2.441 & 0.628 & \bf 0.334 & 8400 & 1600 & \bf 800 \\
\tt LISWET12 & 10002 & 20002 & 50004 & 2.405 & 0.684 & \bf 0.313 & 8400 & 1600 & \bf 800 \\
\tt LISWET2 & 10002 & 20002 & 50004 & 2.012 & 0.717 & \bf 0.283 & 6800 & 1800 & \bf 800 \\
\tt LISWET3 & 10002 & 20002 & 50004 & 1.935 & 0.731 & \bf 0.283 & 6800 & 1800 & \bf 800 \\
\tt LISWET4 & 10002 & 20002 & 50004 & 2.089 & 0.635 & \bf 0.307 & 6800 & 1800 & \bf 800 \\
\tt LISWET5 & 10002 & 20002 & 50004 & 0.907 & 0.397 & \bf 0.212 & 3200 & 1000 & \bf 600 \\
\tt LISWET6 & 10002 & 20002 & 50004 & 2.417 & 0.639 & \bf 0.275 & 8400 & 1600 & \bf 800 \\
\tt LISWET7 & 10002 & 20002 & 50004 & 2.085 & 0.885 & \bf 0.351 & 7200 & 2200 & \bf 1000 \\
\tt LISWET8 & 10002 & 20002 & 50004 & 2.081 & 0.791 & \bf 0.360 & 7200 & 2200 & \bf 1000 \\
\tt LISWET9 & 10002 & 20002 & 50004 & 2.120 & 0.787 & \bf 0.414 & 7200 & 2200 & \bf 1000 \\
\tt LOTSCHD & 12 & 19 & 72 & 0.000 & 0.000 & 0.000 & 400 & 400 & 400 \\
\tt MOSARQP1 & 2500 & 3200 & 8512 & \bf 0.028 & 0.046 & 0.034 & \bf 400 & 600 & \bf 400 \\
\tt MOSARQP2 & 900 & 1500 & 4820 & 0.010 & 0.010 & 0.011 & 200 & 200 & 200 \\
\tt POWELL20 & 10000 & 20000 & 40000 & 136.363 & 283.350 & \bf 0.796 & 462400 & 653200 & \bf 1200 \\
\tt PRIMAL1 & 325 & 410 & 6464 & 0.005 & 0.006 & 0.006 & 200 & 200 & 200 \\
\tt PRIMAL2 & 649 & 745 & 9339 & \bf 0.008 & 0.011 & \bf 0.008 & 200 & 200 & 200 \\
\tt PRIMAL3 & 745 & 856 & 23036 & \bf 0.020 & 0.026 & 0.021 & 200 & 200 & 200 \\
\tt PRIMAL4 & 1489 & 1564 & 19008 & \bf 0.019 & 0.022 & 0.020 & 200 & 200 & 200 \\
\tt PRIMALC1 & 230 & 239 & 2529 & timeout & 0.945 & \bf 0.006 & timeout & 94400 & \bf 600 \\
\tt PRIMALC2 & 231 & 238 & 2078 & timeout & 0.389 & \bf 0.005 & timeout & 45800 & \bf 600 \\
\tt PRIMALC5 & 287 & 295 & 2869 & timeout & \bf 0.005 & \bf 0.004 & timeout & \bf 400 & \bf 400 \\
\tt PRIMALC8 & 520 & 528 & 5199 & timeout & 0.435 & \bf 0.018 & timeout & 21800 & \bf 800 \\
\tt Q25FV47 & 1571 & 2391 & 130523 & \bf 6.124 & timeout & 8.155 & \bf 27600 & timeout & 28200 \\
\tt QADLITTL & 97 & 153 & 637 & 0.004 & 0.004 & 0.004 & 1200 & \bf 1000 & \bf 1000 \\
\tt QAFIRO & 32 & 59 & 124 & 0.000 & 0.000 & 0.000 & 200 & 200 & 200 \\
\tt QBANDM & 472 & 777 & 3023 & 0.228 & \bf 0.044 & 0.049 & 13600 & \bf 2000 & 2200 \\
\tt QBEACONF & 262 & 435 & 3673 & 0.032 & \bf 0.010 & 0.018 & 2600 & \bf 600 & 1000 \\
\tt QBORE3D & 315 & 548 & 1872 & 1.302 & \bf 0.033 & 0.368 & 126200 & \bf 2600 & 29000 \\
\tt QBRANDY & 249 & 469 & 2511 & 0.170 & 0.090 & \bf 0.015 & 14600 & 5600 & \bf 1000 \\
\tt QCAPRI & 353 & 624 & 3852 & 2.041 & 418.003 & \bf 0.088 & 146600 & 22029400 & \bf 4800 \\
\tt QE226 & 282 & 505 & 4721 & 0.557 & 0.147 & \bf 0.077 & 36400 & 7400 & \bf 3400 \\
\tt QETAMACR & 688 & 1088 & 11613 & 0.916 & \bf 0.140 & 0.207 & 10000 & \bf 1200 & 1800 \\
\tt QFFFFF80 & 854 & 1378 & 10635 & \bf 0.362 & 74.270 & 15.281 & \bf 6200 & 1031600 & 201400 \\
\tt QFORPLAN & 421 & 582 & 6112 & \bf 0.009 & timeout & 3.255 & \bf 400 & timeout & 153200 \\
\tt QGFRDXPN & 1092 & 1708 & 3739 & 0.898 & \bf 0.167 & timeout & 43400 & \bf 6600 & timeout \\
\tt QGROW15 & 645 & 945 & 7227 & 463.025 & timeout & \bf 0.121 & 15832000 & timeout & \bf 3400 \\
\tt QGROW22 & 946 & 1386 & 10837 & 29.204 & timeout & \bf 0.116 & 659400 & timeout & \bf 2200 \\
\tt QGROW7 & 301 & 441 & 3597 & 0.536 & \bf 0.036 & timeout & 40600 & \bf 2000 & timeout \\
\tt QISRAEL & 142 & 316 & 3765 & 0.043 & \bf 0.037 & 0.075 & 4800 & \bf 3000 & 6000 \\
\tt QPCBLEND & 83 & 157 & 657 & \bf 0.003 & \bf 0.003 & 0.004 & 1000 & \bf 600 & 800 \\
\tt QPCBOEI1 & 384 & 735 & 4253 & 0.139 & 0.058 & \bf 0.056 & 7000 & 2200 & \bf 1800 \\
\tt QPCBOEI2 & 143 & 309 & 1482 & 0.908 & \bf 0.022 & 0.028 & 148000 & \bf 2200 & 3200 \\
\tt QPCSTAIR & 467 & 823 & 4790 & \bf 0.086 & 29.648 & 0.122 & \bf 3400 & 965200 & 3800 \\
\tt QPILOTNO & 2172 & 3147 & 16105 & \bf 60.362 & timeout & timeout & \bf 411200 & timeout & timeout \\
\tt QPTEST & 2 & 4 & 10 & 0.000 & 0.000 & 0.000 & 200 & 200 & 200 \\
\tt QRECIPE & 180 & 271 & 923 & \bf 0.003 & 0.004 & 0.004 & 600 & 600 & 600 \\
\tt QSC205 & 203 & 408 & 785 & 0.001 & 0.002 & 0.001 & 200 & 200 & 200 \\
\tt QSCAGR25 & 500 & 971 & 2282 & \bf 0.102 & timeout & 0.154 & \bf 8800 & timeout & 9000 \\
\tt QSCAGR7 & 140 & 269 & 602 & 0.036 & 0.435 & \bf 0.005 & 11200 & 86400 & \bf 1000 \\
\tt QSCFXM1 & 457 & 787 & 4456 & \bf 0.278 & 131.058 & 0.872 & \bf 16400 & 5741800 & 41000 \\
\tt QSCFXM2 & 914 & 1574 & 8285 & \bf 1.160 & timeout & 11.558 & \bf 32200 & timeout & 256600 \\
\tt QSCFXM3 & 1371 & 2361 & 11501 & \bf 1.698 & timeout & 2.708 & \bf 30200 & timeout & 40200 \\
\tt QSCORPIO & 358 & 746 & 1842 & timeout & 0.505 & \bf 0.237 & timeout & 40000 & \bf 19400 \\
\tt QSCRS8 & 1169 & 1659 & 4560 & 0.508 & 0.084 & \bf 0.069 & 18200 & 2400 & \bf 2000 \\
\tt QSCSD1 & 760 & 837 & 4584 & 0.023 & 0.017 & \bf 0.013 & 1400 & 800 & \bf 600 \\
\tt QSCSD6 & 1350 & 1497 & 8378 & 0.482 & 0.035 & \bf 0.031 & 16400 & 1000 & \bf 800 \\
\tt QSCSD8 & 2750 & 3147 & 16214 & 0.072 & 0.062 & \bf 0.049 & 1200 & 800 & \bf 600 \\
\tt QSCTAP1 & 480 & 780 & 2442 & timeout & \bf 0.016 & 0.117 & timeout & \bf 1000 & 7600 \\
\tt QSCTAP2 & 1880 & 2970 & 10007 & 0.467 & 0.060 & \bf 0.047 & 8000 & 800 & \bf 600 \\
\tt QSCTAP3 & 2480 & 3960 & 13262 & 0.226 & \bf 0.042 & 0.057 & 2800 & \bf 400 & 600 \\
\tt QSEBA & 1028 & 1543 & 6576 & 0.201 & timeout & \bf 0.151 & 9400 & timeout & \bf 5800 \\
\tt QSHARE1B & 225 & 342 & 1436 & 0.205 & 0.419 & \bf 0.060 & 33800 & 48400 & \bf 6800 \\
\tt QSHARE2B & 79 & 175 & 873 & 0.117 & 1.074 & \bf 0.010 & 36600 & 210800 & \bf 2000 \\
\tt QSHELL & 1775 & 2311 & 74506 & \bf 0.328 & 0.706 & 6.876 & \bf 2600 & 4800 & 41200 \\
\tt QSHIP04L & 2118 & 2520 & 8548 & 0.071 & 0.059 & \bf 0.031 & 1800 & 1200 & \bf 600 \\
\tt QSHIP04S & 1458 & 1860 & 5908 & 0.039 & 0.028 & \bf 0.024 & 1400 & 800 & \bf 600 \\
\tt QSHIP08L & 4283 & 5061 & 86075 & \bf 0.192 & 0.326 & 0.253 & \bf 600 & 800 & \bf 600 \\
\tt QSHIP08S & 2387 & 3165 & 32317 & 0.232 & 0.093 & \bf 0.080 & 2400 & 800 & \bf 600 \\
\tt QSHIP12L & 5427 & 6578 & 144030 & 1.001 & 0.525 & \bf 0.404 & 2000 & 800 & \bf 600 \\
\tt QSHIP12S & 2763 & 3914 & 44705 & 0.186 & \bf 0.056 & 0.093 & 1600 & \bf 400 & 600 \\
\tt QSIERRA & 2036 & 3263 & 9582 & \bf 0.115 & 0.179 & 0.351 & \bf 2000 & 2400 & 4800 \\
\tt QSTAIR & 467 & 823 & 6293 & 2.567 & 317.286 & \bf 0.303 & 89000 & 9359600 & \bf 8200 \\
\tt QSTANDAT & 1075 & 1434 & 5576 & 0.245 & timeout & \bf 0.022 & 10800 & timeout & \bf 800 \\
\tt S268 & 5 & 10 & 55 & 0.000 & 0.000 & 0.000 & 400 & 400 & 400 \\
\tt STADAT1 & 2001 & 6000 & 13998 & timeout & \bf 0.611 & timeout & timeout & \bf 7000 & timeout \\
\tt STADAT2 & 2001 & 6000 & 13998 & timeout & \bf 0.244 & 10.190 & timeout & \bf 3000 & 107800 \\
\tt STADAT3 & 4001 & 12000 & 27998 & timeout & \bf 1.309 & 292.029 & timeout & \bf 7200 & 1489600 \\
\tt STCQP1 & 4097 & 6149 & 66544 & \bf 0.052 & 0.058 & 0.060 & 200 & 200 & 200 \\
\tt STCQP2 & 4097 & 6149 & 66544 & 0.092 & \bf 0.086 & 0.093 & 200 & 200 & 200 \\
\tt TAME & 2 & 3 & 8 & 0.000 & 0.000 & 0.000 & 200 & 200 & 200 \\
\tt UBH1 & 18009 & 30009 & 72012 & 1.106 & \bf 0.463 & 0.711 & 2600 & \bf 800 & 1200 \\
\tt VALUES & 202 & 203 & 7846 & 0.008 & \bf 0.006 & 0.010 & 800 & \bf 600 & 1000 \\
\tt YAO & 2002 & 4002 & 10004 & 224.794 & 7.161 & \bf 4.181 & 4164000 & 111800 & \bf 68000 \\
\tt ZECEVIC2 & 2 & 4 & 7 & 0.000 & 0.000 & 0.000 & 200 & 200 & 200 \\
\midrule
\multicolumn{4}{r}{\bf Problems solved with fewest iterations:} & & & & 15 & 38 & 50 \\
\multicolumn{4}{r}{\bf Problems solved with fastest solve time:} & 31 & 35 & 45 \\
\multicolumn{4}{r}{\bf Total solved before timeout:} & 126 & 125 & 127 \\
    \end{supertabular}
}

\end{document}